%% file: main.tex
\documentclass[letterpaper, 10 pt, conference]{ieeeconf}  % Comment this line out if you need a4paper

\IEEEoverridecommandlockouts                              % This command is only needed if

\usepackage{times}
\usepackage[pdftex]{graphicx}
\usepackage{subfigure}
\usepackage{amsmath,amsopn,amstext,amsfonts}
\usepackage{cancel}
\usepackage[space]{cite}
\usepackage{pdfsync}
\usepackage{balance}
\usepackage{color}
\usepackage{xcolor}
\usepackage{mathtools}
\usepackage{makecell}
\usepackage[linesnumbered,ruled,vlined]{algorithm2e}
\usepackage{bm}

\usepackage{diagbox}
\usepackage{float}
\usepackage{caption}
\usepackage{epstopdf}
\usepackage{pifont}
\usepackage{multirow}
\usepackage{url}
\usepackage{tabularx}

 \usepackage{dblfnote}
 
\usepackage[linkcolor=black,citecolor=black,urlcolor=black,colorlinks=true]{hyperref}

\usepackage{verbatim}
\usepackage{algpseudocode,algorithm2e}
\usepackage{ulem}

\newcommand{\rev}[1]{{\color{black}#1}}
\newcommand{\revsec}[1]{{\color{black}#1}}

\newcommand\blfootnote[1]{%
  \begingroup
  \renewcommand\thefootnote{}\footnote{#1}%
  \addtocounter{footnote}{-1}%
  \endgroup
}

\SetKwInput{KwInput}{Input}                % Set the Input
\SetKwInput{KwOutput}{Output}              % set the Output

\title{\LARGE \bf
EasyHeC: Accurate and Automatic Hand-eye Calibration via Differentiable Rendering and Space Exploration
}

\author{
    Linghao Chen$^{1,2}$
    \quad Yuzhe Qin$^{2}$
    \quad Xiaowei Zhou$^{1}$
    \quad Hao Su$^{2}$
    \\% <-this % stops a space
    $^1$Zhejiang University \quad
    $^2$University of California, San Diego \\
}

\begin{document}

    \maketitle
    \blfootnote{The work is done when Linghao Chen is a visiting graduate student at UC San Diego.
    Corresponding author: Hao Su.}
    \thispagestyle{empty}
    \pagestyle{empty}

%%%%%%%%%%%%%%%%%%%%%%%%%%%%%%%%%%%%%%%%%%%%%%%%%%%%%%%%%%%%%%%%%%%%%%%%%%%%%%%%
    \begin{abstract}
        \input{abstract.tex}
    \end{abstract}

    \input{sections/0_intro_new.tex}
    \input{sections/1_related.tex}

    \input{sections/3_method.tex}
    \input{sections/4_experiments.tex}
    \input{sections/5_limitation.tex}
    \input{sections/6_conclusions.tex}

    \normalem
    \bibliography{main}
    \bibliographystyle{IEEEtran}

\end{document}

%% file: abstract.tex
Hand-eye calibration is a critical task in robotics, as it directly affects the efficacy of critical operations such as manipulation and grasping.
Traditional methods for achieving this objective necessitate the careful design of joint poses and the use of specialized calibration markers, while most recent learning-based approaches using solely pose regression are limited in their abilities to diagnose inaccuracies.
In this work, we introduce a new approach to hand-eye calibration called EasyHeC, \rev{which is markerless, white-box, and delivers superior accuracy and robustness.}
We propose to use two key technologies: differentiable rendering-based camera pose optimization and consistency-based joint space exploration, which enables accurate end-to-end optimization of the calibration process and eliminates the need for the laborious manual design of robot joint poses. Our evaluation demonstrates superior performance in synthetic and real-world datasets, enhancing downstream manipulation tasks by providing precise camera poses for locating and interacting with objects.
The code is available at the project page: \url{https://ootts.github.io/easyhec}.

%% file: sections/0_intro_new.tex
\section{INTRODUCTION}

Hand-eye calibration, a fundamental problem in robotics with a long history, addresses the problem of determining the transformation between a camera and a robot.
It aligns the two most important modalities for robot sensing: proprioception and perception.
Thus, hand-eye calibration serves as a crucial first step for various visual robotics applications.

Traditional calibration methods rely on markers, such as ArUco tags~\cite{garrido2014automatic} or checkerboard patterns~\cite{ilonen2011robust,tsai1989new}, to establish a connection between the robot and camera frames. Despite being grounded in robust theory, numerous practical challenges arise when implementing these methods in real-world robotic systems. Research indicates that the accuracy of traditional approaches is considerably influenced by factors such as marker fixture quality, marker localization accuracy, and the choice of joint poses. Crucially, varying camera placements for distinct tasks require the creation of different joint pose trajectories, a task that is highly demanding for humans. While industry users may achieve satisfactory accuracy through extensive tuning for specific tasks, traditional marker-based methods fall short in the quest to develop versatile robots with high accuracy and minimal human intervention in daily environments. Consequently, there is an urgent demand for an accurate, automated hand-eye calibration solution that is both easy to implement and user-friendly. This necessity is especially pertinent given the growing application of computer vision algorithms across a broad spectrum of robotics tasks.

\input{figures/teaser.tex}

More recently, efforts have been made to develop marker-less hand-eye calibration techniques~\cite{labbe2021single,sefercik2023learning,bahadir2022deep}, utilizing learning-based methods to minimize the cost and labor associated with calibration. However, the performance of these approaches largely depends on the scale and quality of their training data. Both traditional and learning-based methods exhibit uneven hand positioning error distribution in the camera space across the robot's configurations. Greater accuracy is achieved for configurations closely resembling the sampled ones, while larger discrepancies are observed in areas of the configuration space that remain unexplored during calibration. Additionally, most learning-based methods lack transparency due to their reliance on direct regression, making it challenging to diagnose calibration errors in real-world applications.

In this work, we aim to simplify hand-eye calibration for robotics researchers.
To this end, we propose a system called EasyHeC, which offers several key features:
\begin{itemize}
    \item No need for markers to be attached to the robot;
    \item Easy diagnosis with a white-box design;
    \item Elimination of laborious manual design of robot joint poses for calibration;
    \item \rev{High accuracy and robustness};
\end{itemize}
To achieve our goal with EasyHeC, we propose the use of two key technologies: (i) differentiable rendering-based camera pose optimization and (ii) consistency-based joint space exploration.

First, we substitute the widely used objective function based on transformation equivalence, such as $AX=XB$~\cite{tsai1989new, park1994robot, daniilidis1999hand}, with a loss function grounded in per-pixel segmentation of the robot arm.
We assume that the 3D model of the robot arm is readily available, which is a reasonable assumption as most commonly used arms can be found online.
In our formulation, the transformation between the camera and the robot base frames is optimized by minimizing the discrepancy between the projection of the 3D arm model and the observed segmentation mask.
This white-box approach establishes a direct connection between the camera observations and camera poses, enabling end-to-end optimization of the calibration process and rendering our system highly diagnosable.
Moreover, by leveraging the shape of the robot arm itself, we eliminate the need for calibration markers and reduce the financial cost of the hand-eye calibration process.
As the robot arm is considerably larger than the markers attached to it, the arm's mask offers substantially more evidence for estimating the transformation than traditional marker points.

Second, we develop a consistency-based joint space exploration module that facilitates the selection of informative joint poses.
This module identifies the highest informative robot joint poses by evaluating the consistency with a set of camera pose candidates.
This automated selection process eliminates the need for manual design of the robot joint poses.
Moreover, since this process is performed in simulation, it is free of cost and can be performed many times to improve the accuracy of the calibration.

We evaluated our method on several synthetic and real-world datasets and provide ablation studies of the different design choices of the proposed system.
When evaluated on synthetic datasets, EasyHeC \textbf{outperforms a traditional-based baseline by 4 times in accuracy} with a single view, and achieves an accuracy of around $0.2$ cm with 5 views.
Upon evaluation using a publicly available real-world dataset, EasyHeC consistently demonstrates superior performance across all metrics. Furthermore, we show that our proposed hand-eye calibration system can enhance downstream manipulation tasks by providing precise camera poses for locating and interacting with objects. 
Our system is easy to deploy and user-friendly, and we have made it open-source for the benefit of the robotics community.

%% file: figures/teaser.tex
\begin{figure}
{\centering
\resizebox{0.5\textwidth}{!}{
    \includegraphics[width=15cm]{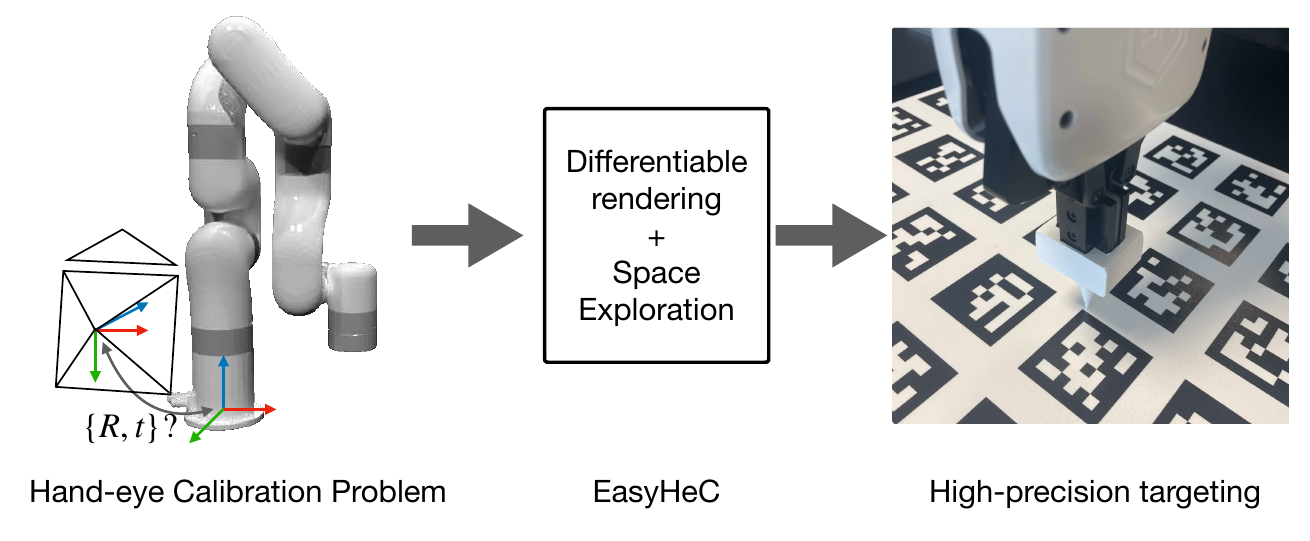}}
}
    \vspace{-1em}
    \caption{\textbf{We propose EasyHeC for hand-eye calibration.}
    Left: Hand-eye calibration problem. We aim to estimate the relative pose between the camera and the robot's base.
    Middle: The main techniques of our approach EasyHeC are differentiable rendering (\ref{subsec:diff-rendering}) and space exploration (\ref{subsec:space-exploration}).
    Right: High-precision targeting with a real robot arm.
    }
   \vspace{-3em}
    \label{fig:teaser}
\end{figure}

%% file: sections/1_related.tex
\section{RELATED WORK}\label{sec:related-work}

\medskip
\noindent\textbf{Traditional Hand-eye Calibration.}
The Hand-eye Calibration (HeC) problem encompasses two settings: (i) \rev{eye-to-hand}, where ``eye", the camera is fixed to the world; and (ii) \rev{eye-in-hand}, where the camera is fixed to ``hand", commonly the end-effector.
Traditional methods for HeC involve the explicit tracking of a specific marker~\cite{garrido2014automatic}.
Depending on the HeC setting, the marker is attached to the robot end-effector for \rev{eye-to-hand} or fixed to the ground for \rev{eye-in-hand}. The camera pose is then obtained by solving a linear system~\cite{park1994robot,ilonen2011robust,tsai1989new,daniilidis1999hand}. However, these methods necessitate complete visibility of the marker, which complicates the design of a calibration trajectory. To address this limitation, marker-free methods~\cite{andreff2001robot,heller2011structure,zhi2017simultaneous} have been proposed, which use Structure-from-Motion (SfM) to estimate camera motion in the eye-in-hand setting. However, such methods rely on rich visual features for correspondence and are not suitable for tracking robot motion in the eye-to-hand setting. Unlike the methods mentioned above, our work addresses the eye-to-hand HeC problem without the use of markers. Instead of solving a linear system, our approach formulates HeC as an iterative optimization problem by minimizing the per-pixel discrepancy of the robot mask.

\noindent\textbf{Learning-based Hand-eye Calibration.}
Several recent methods have proposed learning-based approaches to solve the HeC problem.
For example, DREAM~\cite{lee2020dream} uses a deep neural network to predict robot joint keypoints and then recovers camera poses using the Perspective-n-point (PnP) algorithm~\cite{lepetit2009epnp}.
\rev{Labbé et al.}~\cite{labbe2021single} uses a neural network as a refinement module to iteratively predict robot delta states from rendered images and observations.
However, these regression-based methods fail to establish a direct link between the camera pose and observation, thereby limiting their practical diagnostic ability.
\rev{Valassakis et al.}~\cite{valassakis2022learning} directly regresses camera pose from images but lacks robustness in practical scenarios.
\rev{Sefercik et al.}~\cite{sefercik2023learning} uses neural networks for end-effector segmentation and the iterative closest point (ICP) algorithm~\cite{besl1992icp} for camera pose estimation, but relies on depth sensors and may not work well without high-quality depth data.
\rev{Lu et al.~\cite{lu2022pose} proposed to first detect an optimal set of keypoints through an iterative process and then utilize PnP to solve the transformation between the camera and the robot base.
The main difference between our method and theirs is that we use a differentiable renderer to optimize the poses without the need for defining keypoints.}
\rev{Bahadir et al.}~\cite{bahadir2022deep} tracks a reference point on the robot and regresses the camera pose.
\rev{A recent work~\cite{lu2023image} proposed a similar idea with us by using differentiable rendering to solve the hand-eye calibration problem. However, there are two main differences between our method and theirs.}
\rev{First, they attempt to optimize the camera pose and robot shape simultaneously using a single image. This approach introduces ambiguity into the optimization process, as minor adjustments to the robot's shape or the camera's proximity to the robot can result in similar images. Ultimately, they were unable to achieve high accuracy. In contrast, we make a reasonable assumption that the robot shape is already known, and we focus solely on the camera pose as the variable. This assumption significantly enhances the accuracy of our method.
Second, they treated the robot arm as a static rigid body, failing to fully utilize the articulated structure of the robot. In contrast, we use the space exploration module to greatly enhance the accuracy of the calibration results while maintaining a fully automatic process.}
\input{figures/pipeline.tex}

\noindent\textbf{\rev{Next best view selection.}}
\rev{The problem of selecting the next best view has found extensive application in the field of robotics.
Zhang et al.~\cite{zhang2023active}  introduced a discrete view quality field to find the most representative camera view.
Yang et al.~\cite{yang2023next} proposed estimating calibration uncertainty as a means to determine the next viewpoint that offers the highest information gain.
We employ a similar approach in our work by selecting the next best joint pose to improve calibration accuracy. However, their approaches are designed for the eye-in-hand configuration using a marker, whereas our method is designed for the eye-to-hand setup without any markers.
In addition, our method calculates the camera pose uncertainty in the mask space, which has not been explored in existing works.}

%% file: figures/pipeline.tex
\begin{figure*}[ht]
    \centering
    \includegraphics[width=\linewidth]{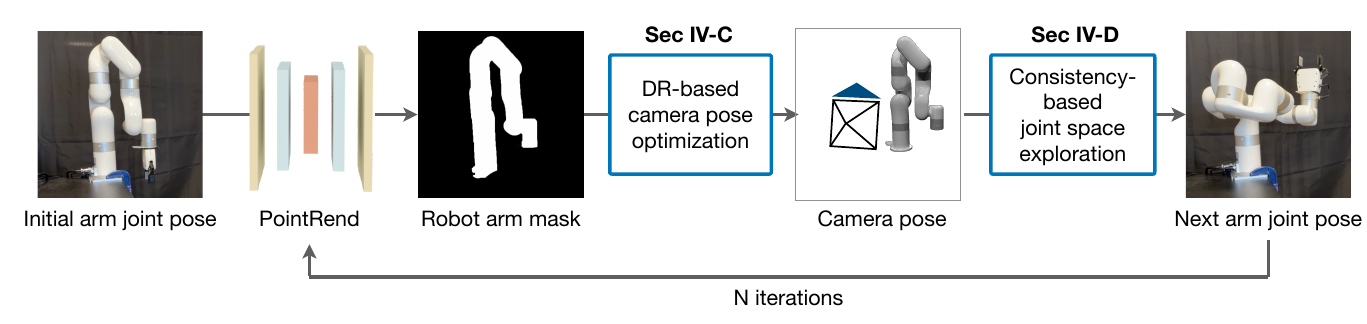}
    \vspace{-1em}
    \caption{
        \textbf{EasyHeC architecture.}
        The architecture is composed of three main components that are executed iteratively.
        First, the RGB images are fed into a mask prediction network (PointRend) to acquire the observed robot arm mask.
        Second, a differentiable rendering (DR)-based camera pose optimization module is used to produce the camera pose by minimizing the difference between the rendered mask and the observed mask.
        Next, the robot arm is driven to the next arm joint pose generated by the consistency-based joint space exploration module.
        We then repeat the above steps for $N$ iterations.
    }
    % \vspace{-2em}
    \label{fig:pipeline}
\end{figure*}

%% file: sections/3_method.tex
\section{METHOD}\label{sec:method}

\subsection{\rev{Problem formulation}}\label{sec:problem_formulation}
Given a robot arm with a known 3D model and a camera with known intrinsic parameters, the objective is to solve the eye-to-hand calibration problem, which is to estimate the transformation between the camera $c$ and the robot base $b$.
In this paper, the transformation is represented as $T_{cb} \in \mathit{SE}$(3) that satisfies: $P^{c}=T_{cb} P^{b}$,
where $P^{c}$ and $P^{b}$ are the 3D points in the camera coordinate system and the robot base coordinate system, respectively.

\subsection{Overview}\label{subsec:overview}

As shown in Fig.~\ref{fig:pipeline}, we optimize the camera pose in an iterative manner.
In each iteration, given the image of the robot arm at a specific joint pose, we first predict the observed mask using PointRend~\cite{kirillov2020pointrend}.
Following this, we utilize differentiable rendering (Sec.~\ref{subsec:diff-rendering}) to refine the camera pose by minimizing the discrepancy between the rendered mask and the observed mask.
Next, to optimize the $T_{cb}$ more efficiently, we utilize
the consistency-based joint space exploration module (Sec.~\ref{subsec:space-exploration}) to select the most informative joint pose for the next iteration.
The robot arm is then driven to this informative joint pose and the proposed method proceeds to the next iteration.
Note that in later iterations, the camera pose is optimized with the masks at all the observed joint poses.

\subsection{Differentiable rendering-based camera pose estimation}\label{subsec:diff-rendering}
Different from the equation $AX=XB$~\cite{tsai1989new} that is widely used in marker-based hand-eye calibration, our approach employs a novel objective function that operates at the pixel level.
Assuming that the 3D model of the robot arm is known, we can obtain the camera pose by minimizing the difference between the rendered robot arm and the camera observation, which is similar to the render-and-compare technique used in~\cite{kundu20183d}.
Unlike marker-based hand-eye calibration methods and regression-based methods, the pixel-wise objective function provides a direct connection between the camera pose and the observed image. This feature makes our approach considerably more diagnosable.

Specifically, we use the objective function to minimize the difference between the rendered mask and the observed mask:
\begin{equation}
    L(T_{cb})=(\min(1,\sum_l \pi(T_{cb} T_{bl} l))-M)^2,
\end{equation}
where $\pi$ is the differentiable mask renderer, $l$ represents the 3D shape of a robot link, $M$ is the mask of the robot, $T_{cb}$ is the transformation between the robot base $b$ and the camera $c$, and $T_{bl}$ is the transformation between the link $l$ and the robot base $b$, which can be obtained through forward kinematics with joint poses.
\rev{Since we generate masks for each individual link and combine them to form the mask for the entire robot arm, we utilize the $\min$ function to limit the sum of all the link masks to a maximum value of 1.}

As the optimization target $T_{cb}$ belongs to the $\mathit{SE}(3)$ space, we transform the transformation matrix to the Lie algebra space~\cite{gabay1982minimizing} and then perform the optimization using the following objective function:
\begin{equation}
    L(\xi_{cb})=(\min(1,\sum_l \pi(\exp(\xi_{cb}) T_{bl} l))-M)^2,\label{eq:dr}
\end{equation}
The above loss function is fully differentiable, we could use the gradient descent method with PyTorch~\cite{paszke2017automatic} Autograd to perform the optimization.

\medskip\noindent\textbf{Mask prediction.}
To obtain the robot arm mask $M$ as the supervision for the differentiable rendering-based optimization, we leverage the off-the-shelf segmentation network PointRend~\cite{kirillov2020pointrend} to predict the mask from the RGB image.
We first train the PointRend model on synthetic data generated by the SAPIEN~\cite{Xiang_2020_SAPIEN} robot simulator and then fine-tune it on real data to enhance its generalization ability.

\medskip\noindent\textbf{Camera pose initialization.}
In our work, we adopt the PVNet~\cite{peng2019pvnet} to perform the pose initialization.
Given an RGB image of the robot arm with the initial joint pose, PVNet estimates the pre-defined 2D keypoint locations through pixel-wise voting and subsequently solves the transformation using PnP~\cite{lepetit2009epnp}.

\subsection{Consistency-based joint space exploration}\label{subsec:space-exploration}
In this subsection, a novel approach is presented that alternates between identifying the most informative supplementary joint pose for camera pose estimation and recalculating the camera pose by incorporating this new joint pose.

Recall that, our camera pose estimation is achieved by solving Eq.~\ref{eq:dr}, which aims to align the mask of the projected 3D model of the arm with the mask of the observed arm.
We can rewrite the objective function in a more simplified manner:
\begin{equation}
    L(t; J)=\left\|\pi(J, t)-M_J\right\|^2,\label{dr2}
\end{equation}
where $\pi$ is a mask renderer, $J$ is a joint pose, $t$ is a camera pose, $M_J$ is the observed robot arm mask at joint pose $J$.
This 2D matching constraint is often insufficient in constraining the solution space of the 6D camera pose given limited joint pose observations.
For example, accurately estimating the camera's distance to the robot base can be challenging with only one joint pose observation.
In other words, there might be a set of reasonable camera pose candidates $T$ that satisfies $L(t,J) < \epsilon$, where $\epsilon$ is a small value.

The goal is to shrink $T$ as much as possible by obtaining a new $T' \subseteq T$ as the solution set of $L(t; J')<\epsilon$ with a new joint pose $J'$.
The next question is, how to choose $J'$ so that $T$ can be shrunk as soon as possible?
Equivalently, how to choose $J'$ so that $L(t,J')>\epsilon$ for as many $t\in T$ as possible?

Our approach is to sample a representative set of $\left\{t_i\right\}_{i=1}^K \in T$, and based upon which, we use $\sum_i L(t_i,J')$ as the surrogate to help us to choose $J'$.
In fact, it can be proved that:
\begin{equation}
    \sum_i L(t_i,J') \geq \operatorname{Var}(m_1, m_2, \ldots, m_K),
\end{equation}
where $m_i=\pi (J',t_i)$.
This inequality then tells us that, if we choose $J'$ to maximize the variance of the masks, we can shrink $T$ as soon as possible.
\input{algorithms/space_exploration}

Based on this observation, we develop a joint space exploration strategy that operates solely in simulation, allowing us to generate informative joint poses before executing motion commands on the robot.
The algorithm is summarized in Alg.\ref{alg:space_exploration}, where we first sample a large number of joint poses in the simulator and then evaluate the variance of the rendered masks over the camera pose candidates.
Then, we move the robot to the joint pose with the highest variance and perform the next iteration of DR-based optimization. Unlike traditional marker-based methods that necessitate the manual design of joint poses, our approach is fully automatic and requires no human intervention.

In the subsequent DR-based optimization iterations, we include all previously observed masks rather than using only the most recent observation. The final optimization objective is defined as follows:
\begin{equation}
    L(\xi_{cb})=\dfrac{1}{K}\sum_{i=1}^K(\min(1,\sum_l \pi(\exp(\xi_{cb}) T_{bl,i} l))-M_i)^2,
\end{equation}
where $K=\{1,2,\dots,N\}$ is the number of images, $T_{bl,i}$ is the transformation between the link $l$ and the robot base $b$ at the $i$-th joint pose, and $M_{i}$ is the mask of the robot in the $i$-th image.

\subsection{Implementation details}\label{subsec:impl-details}
In our implementation, we employ nvdiffrast~\cite{laine2020modular} as the differentiable mask renderer. During the optimization process, we utilize the Adam optimizer with a learning rate of 3e-3 and perform 1000 steps.
For the space exploration module, we randomly choose 50 camera poses from the range between step 200 and 1000 as camera pose candidates. In practice, the number of sampled camera pose candidates and the range may vary depending on the number of optimization iterations.
Next, we randomly sample 2000 joint poses and filter out any invalid poses that result in self-intersection of the robot arm, links exceeding a pre-defined distance threshold from the robot arm base, or collisions with the surrounding environment. Subsequently, we identify the joint pose with the largest variance as the next joint pose to explore.
The stopping criterion can be set based on the desired level of accuracy.
Typically, the calibration error will converge after enough iterations.
In real-world scenarios where the calibration error is unavailable, the change in variance between two iterations of space exploration can be used as a proxy to determine the stopping criterion.

%% file: algorithms/space_exploration.tex
\begin{algorithm}
    \DontPrintSemicolon
    \KwInput{
        Camera pose candidates $T = \left\{t_i\right\}_{i=1}^K$; \\
        Joint pose sampling number $N_{s}$; \\
    }
    \KwOutput{The next joint pose $J'$}

    Randomly sample $N_{s}$ joint poses $J_{1}, J_{2}, \dots, J_{N_{s}}$. \\

    \For{each joint pose $J_i$}{
        \For{each camera pose $t_k$}{
            // Render the robot mask with the joint pose $J_i$ and the camera pose $t_k$. \\
            $m_{ij}=\pi(J_i,t_k)$
        }
        $V_{i} = \text{Var}(m_{i1},m_{i2},...,m_{iK})$
    }
    $J'=J_{argmax(V)}$
    \caption{Joint space exploration.}\label{alg:space_exploration}
\end{algorithm}

%% file: sections/4_experiments.tex
\section{EXPERIMENTS}

\subsection{Evaluation on synthetic dataset}\label{subsec:eval-on-syn}

\medskip\noindent\textbf{Data generation.}
In our first experiment, we use SAPIEN~\cite{Xiang_2020_SAPIEN} to synthesize 100 photo-realistic scenes with different camera poses using an xArm robot arm and compare our method with the traditional marker-based method.
We use 20 manually designed joint poses for each scene for the marker-based method.

\input{figures/xarm_syn_example.tex}

Our method only required the image with the initial joint pose and the subsequent views are generated online.
An example of the synthetic dataset is shown in Fig.~\ref{fig:xarm_syn_example}, where we attach a chessboard marker on the robot arm for the marker-based method.
The size of the marker is 4cm$\times$5cm, with each grid being 1cm$\times$1cm.
We rendered 10,000 images for the PointRend~\cite{kirillov2020pointrend} and another 10,000 images for the PVNet~\cite{peng2019pvnet} training to obtain the observed segmentation mask and initial camera pose for our method.

\medskip\noindent\textbf{Evaluation of hand-eye calibration.}
We perform evaluation based on the translation error and rotation error between the ground-truth camera poses and the solved camera poses.
Note that the manually designed joint poses do not always ensure that the marker is visible in all scenes, especially if there is occlusion.
To ensure fairness, we also consider randomly sampling a large number of joint poses for each scene and selecting only those where the marker is visible.
However, we find that only 1.9\% of the sampled joint poses result in clear marker visibility, which makes this approach impractical.
Thus, we use the same joint poses for all scenes but only analyze those scenes where more than 6 markers are detected.

The results are shown in Tab.~\ref{tab:xarm_synthetic_rotation} and Tab.~\ref{tab:xarm_synthetic_translation}, where our method outperforms the marker-based method~\cite{tsai1989new} by a large margin.
Specifically, the average rotation and translation errors of the marker-based method are 0.9 degrees and 2.1 cm, respectively.
For the proposed method, the evaluation is performed using different numbers of views ranging from 1 to 5.
The errors consistently decrease with an increase in the number of views.
Specifically, when there is only 1 view, the error of our method is 0.32 degrees and 0.5 cm, whereas, with 5 views, the error decreases to 0.08 degrees and 0.2 cm.
\input{tables/xarm_synthetic_rotation}
\input{tables/xarm_synthetic_translation}
DREAM~\cite{lee2020dream} uses a network to detect keypoints on the robot arm and uses the PnP algorithm to estimate the camera pose.
Since the keypoints are not discriminative, the network often fails to accurately detect the keypoints.
Moreover, detecting keypoints is not resilient to occlusion and truncation, thus the method produces a low-accuracy result, especially for the rotation error.
Even if the performance of the DREAM method is enhanced by employing multiple views, it still produces worse performance compared to our method.
In contrast, our method eliminates the need for any keypoints and demonstrates robustness in the presence of occlusion and truncation.
\subsection{Evaluation on Real-world Baxter dataset}\label{subsec:eval-on-real-baxter}

\medskip\noindent\textbf{Dataset introduction.}
In this section, we conduct experiments on the real-world Baxter dataset~\cite{lu2022pose} and compare the efficacy of our method against state-of-the-art learning-based methods.
The Baxter dataset contains 100 images of a real Baxter robot, with a single camera pose and 20 distinct joint poses.
In light of the single camera pose, we manually initialize the camera pose for simplicity.

\medskip\noindent\textbf{Implementation details and evaluation metrics.}
To train the segmentation network, we train the PointRend first on the synthetic DREAM dataset~\cite{lee2020dream} with domain randomization and then fine-tune on some realistic images rendered with SAPIEN~\cite{Xiang_2020_SAPIEN}.
The experimental results are presented in Tab.\ref{tab:baxter_main_2d} and Tab.\ref{tab:baxter_main_3d}.
\input{figures/baxter_result_example.tex}
\input{tables/baxter_main_2d}
\input{tables/baxter_main_3d}
\rev{We report the 2D and 3D percentage of correct keypoints (PCK), which is the percentage of 2D/3D keypoints that are within a certain threshold distance from the ground truth.}
We also provide some visualizations of the results in Fig.~\ref{fig:baxter_result_example}.
In addition to the thresholds employed in ~\cite{lu2023image,lu2022pose}, we report results with smaller thresholds to further elucidate the accuracy of our method since our method nearly reaches 100\% at the thresholds used in previous works.
Following the conventions in~\cite{lu2023image,lu2022pose}, we present the evaluation results on the training images in Tab.~\ref{tab:baxter_main_2d} and Tab.~\ref{tab:baxter_main_3d}, while in Fig.~\ref{fig:baxter_ablation}, we present the results evaluated on the entire dataset since it is more representative to evaluate the performance at other joint poses.

\medskip\noindent\textbf{Evaluation of hand-eye calibration.}
For our method, we perform evaluations with 1 to 3 views.
The results demonstrate that our method outperforms the previous state-of-the-art methods by a large margin,
demonstrating the effectiveness and robustness of our method.

\subsection{Evaluation on the real-world xArm data}\label{subsec:eval-on-real-Xarm}

To assess the effectiveness of our proposed hand-eye calibration method in real-world scenarios, we conduct a high-precision targeting experiment using an \rev{ArUco} marker board and a pointer attached to the robot's end effector.

We detect the corners of the marker board using OpenCV~\cite{bradski2000opencv} and transform their positions to the robot arm base coordinate system via our hand-eye calibration results.
\input{tables/xarm_real}
Next, we employ the pointer to tip the marker board corners and manually measure the distance between the pointer and the actual corner position.
A representative illustration is visualized in Fig.~\ref{fig:teaser}.

The results and comparison with DREAM are presented in Tab.~\ref{tab:xarm_real}.
\revsec{Since DREAM estimates 2D keypoints and uses the PnP algorithm to estimate the camera pose, it is not robust to occlusion and truncation and suffers from the sim-to-real domain gap.
In contrast, our method uses dense pixel-wise error as the objective function, thus it is inherently robust to truncation and self-occlusion.}
\revsec{Although the manual measurement is not highly reliable, it offers a preliminary indication of the system's accuracy and convincingly demonstrates the effectiveness of our approach.}
To further improve the accuracy, we can use a more precise segmentation network such as SAM~\cite{kirillov2023segany} to reach a higher accuracy as shown in Tab.~\ref{tab:xarm_real}.
\revsec{In practice, the selection between these two segmentation methods depends on specific needs.
PointRend involves a preliminary pre-training but offers a fully automated calibration process, whereas SAM mandates simple mask labeling but requires human intervention in the process.}

\subsection{Ablation studies}\label{subsec:ablation}

\medskip\noindent\textbf{Joint space exploration.}
To demonstrate the effectiveness of the joint space exploration module, we conduct an ablation study by comparing it with the random selection approach.
The experiments were performed on the xArm synthetic dataset and the Baxter dataset.
For the xArm synthetic dataset, the joint pose selection is from the sampled joint pose, whereas for the Baxter dataset, the joint pose selection is from the joint poses provided by the dataset.
\rev{The results are shown in Tab.~\ref{tab:xarm_synthetic_rotation}, Tab.~\ref{tab:xarm_synthetic_translation}, and Fig.~\ref{fig:baxter_ablation}.}
The results on the xArm synthetic dataset show that using space exploration to select the next joint pose consistently provides faster convergence speed and produces smaller errors than random selection.
The results on the Baxter dataset show that the space exploration-based approach produces higher PCK than the random selection approach, especially for the PCK2D.

\input{figures/baxter_ablation.tex}
\input{figures/xarm_syn_ablation_trans.tex}

\medskip\noindent\textbf{Different number of sampled joint poses.}
To assess the influence of varying numbers of joint poses on our method's performance in the joint space exploration process, we conducted experiments on the xArm synthetic dataset by sampling different numbers of joint poses. The outcomes of these experiments are presented in Fig.~\ref{fig:xarm_syn_ablation_trans}, where it is demonstrated that the errors decrease as the number of sampled joint poses increases, particularly when the number of joint poses is below 1000. As soon as the number of joint poses exceeds 1000, however, the errors become virtually unchanged.
Therefore, we set the number of sampled joint poses to 2000 in other experiments.
\input{tables/runtime}

\medskip\noindent\textbf{Different number of sampled camera candidates.}
To verify the influence of different numbers of sampled camera candidates in the joint space exploration process on the performance of our method, we conduct an ablation study on the xArm synthetic dataset by varying the number of sampled camera candidates.
\rev{The results in Fig.~\ref{fig:xarm_syn_ablation_trans} show} that the errors are roughly the same as the number of sampled camera candidates varying.
This indicates that the number of sampled camera candidates does not affect the performance of our method, thus we can use a small number of camera pose candidates in practice to save computation time.
In other experiments, we set the number of sampled camera candidates to 50.

\subsection{\rev{Running time}}\label{subsec:running-time}
\rev{In general, the total runtime of the system consists of three main components: DR-based optimization, space exploration, and the robot's movement between joint poses in real-world applications.
In practice, the runtime depends on the number of optimization steps, space exploration iterations, camera pose candidates, and sampled joint poses.
In real experiments, the runtime also depends on factors such as the robot's speed, the distance between the joint poses, and the planned trajectory.}
\revsec{The runtime of our method is presented in Tab.~\ref{tab:runtime}. This runtime is considered acceptable when compared to time-consuming marker-based calibrations~\cite{tsai1989new}, which can take several hours to finetune.
It's important to note that once the calibration is completed, there is no need to repeat it unless the camera or the robot is replaced.}

\subsection{Real-world applications}\label{subsec:real-world-apps}
EasyHeC is useful in a variety of real-world applications~\cite{jia2023chain,chen2023perceiving}.
In this section, we present some real-world applications to exemplify the efficacy of our proposed approach.
Suppose we intend to employ a state-based policy with solely visual inputs, then it becomes imperative to convert object states from the camera coordinate system to the robot base coordinate system.
In this case, the proposed EasyHeC could be leveraged to perform this transformation.
In our experiment, we choose CoTPC~\cite{jia2023chain} as the state-based algorithm, which takes states as input and outputs the next robot action.
We perform two experiments using the CoTPC algorithm: stacking cube and peg insertion.
Specifically, we use PVNet~\cite{peng2019pvnet} to estimate the object poses in the camera coordinate system and then use the hand-eye calibration results to transform the object poses from the camera coordinate system to the base coordinate system as the input states to the CoTPC network.
Fig.\ref{fig:cotpc} shows the screenshots captured during the execution process, which demonstrate that the cubes are accurately aligned and the peg is successfully inserted, thereby affirming the effectiveness of our method in state-based algorithms.

\input{figures/cotpc.tex}

%% file: figures/xarm_syn_example.tex
\begin{figure}
    \centering
    \resizebox{1.0 \columnwidth}{!}{
        \begin{tabular}{cc}
            \includegraphics[width=0.5\linewidth,trim={0cm 0cm 0cm 0cm},clip]{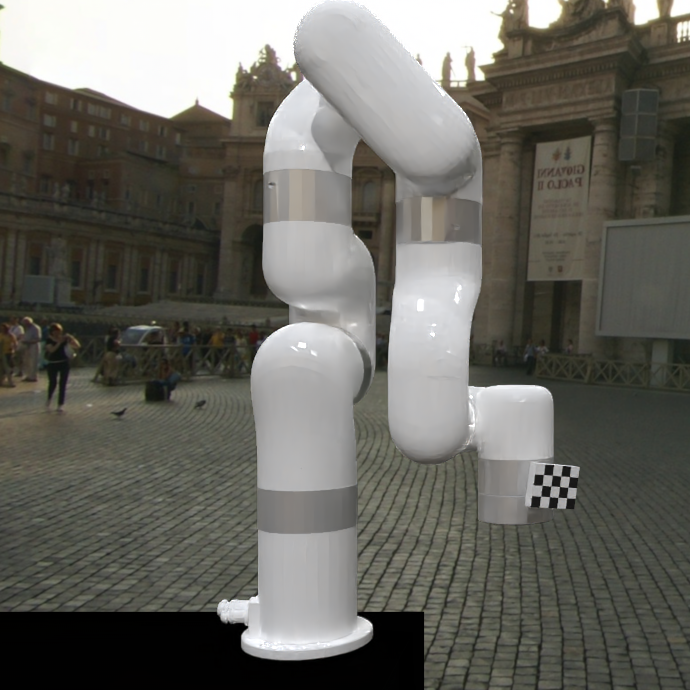} &
            \includegraphics[width=0.5\linewidth,trim={0cm 0cm 0cm 0cm},clip]{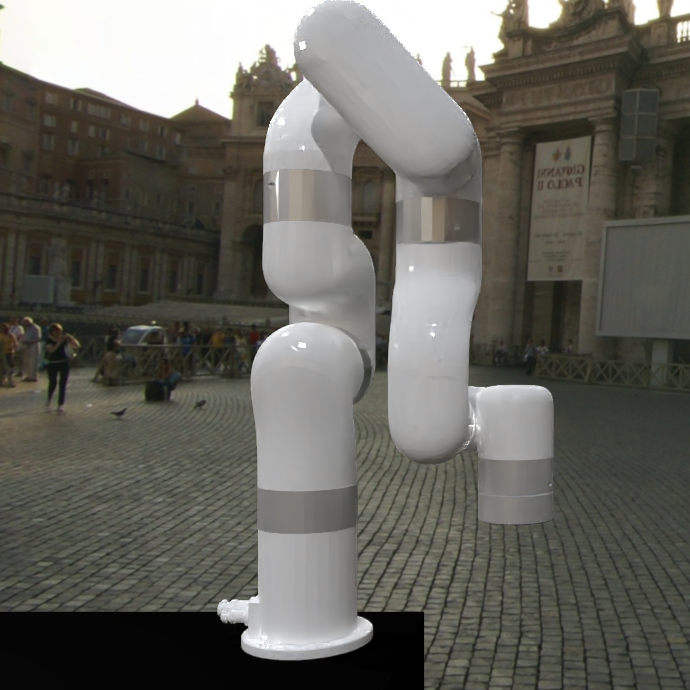}\\
            (a) & (b) \\
        \end{tabular}
    }
    \caption{\textbf{\rev{Example images in the xArm synthetic dataset.}}
    (a) for marker-based method and (b) for our method.
    }
    \vspace{-4em}
    \label{fig:xarm_syn_example}
\end{figure}

%% file: tables/xarm_synthetic_rotation.tex
\begin{table}[ht]
    \centering
    \renewcommand{\arraystretch}{1.1}
    \resizebox{0.45\textwidth}{!}{
        \begin{tabular}{cccccc}
            \hline
            Method      & 1     & 2     & 3     & 4     & 5     \\
            \Xhline{3\arrayrulewidth}
            Marker-based & 0.870 & - & - & - & - \\
            \hline
            \rev{DREAM~\cite{lee2020dream}} & \rev{1.924} & \rev{1.240} & \rev{0.981} & \rev{0.764} & \rev{0.704} \\
            \hline
            Ours (Rand) & 0.318 & 0.214 & 0.151 & 0.112 & 0.098 \\
            \hline
            Ours (SE)   & \textbf{0.322} & \textbf{0.128} & \textbf{0.109} & \textbf{0.097} & \textbf{0.081} \\
            \hline
            \Xhline{3\arrayrulewidth}
        \end{tabular}
    }
    \caption{
        \textbf{Rotation error evaluation results on the xArm synthetic dataset.}
        For our method, we report errors in degrees with the number of views from 1 to 5.
        Ours (SE) and Ours (Rand) represent our method with the next joint configuration generation using consistency-based space exploration and random sampling, respectively. Note that the Marker-based method utilizes 20 views to compute hand-eye calibration results.    }
    \label{tab:xarm_synthetic_rotation}
\end{table}

%% file: tables/xarm_synthetic_translation.tex
\begin{table}[ht]
    \centering
    \renewcommand{\arraystretch}{1.1}
    \resizebox{0.45\textwidth}{!}{
        \begin{tabular}{cccccc}
            \hline
            Method      & 1     & 2     & 3     & 4     & 5     \\
            \Xhline{3\arrayrulewidth}
            Marker-based & 2.000 & - & - & - & - \\
            \hline
            \rev{DREAM} & \rev{0.529} & \rev{0.473} & \rev{0.374} & \rev{0.347} & \rev{0.303} \\
            \hline
            Ours (Rand) & 0.496 & 0.388 & 0.361 & 0.322 & 0.312 \\
            \hline
            Ours (SE) & \textbf{0.488} & \textbf{0.298} & \textbf{0.252} & \textbf{0.206} & \textbf{0.206} \\
            \Xhline{3\arrayrulewidth}
        \end{tabular}
    }
%     \vspace{-0.3cm}
    \caption{
        \textbf{Translation error evaluation on the xArm synthetic dataset.}
        We report errors in centimeters with the number of views from 1 to 5.
    }
    \vspace{-5em}
    \label{tab:xarm_synthetic_translation}
\end{table}

%% file: figures/baxter_result_example.tex
\begin{figure}[ht]
    \centering
    \resizebox{1.0 \columnwidth}{!}{
        \begin{tabular}{cc}
            \includegraphics[width=0.5\linewidth,trim={0cm 0cm 0cm 0cm},clip]{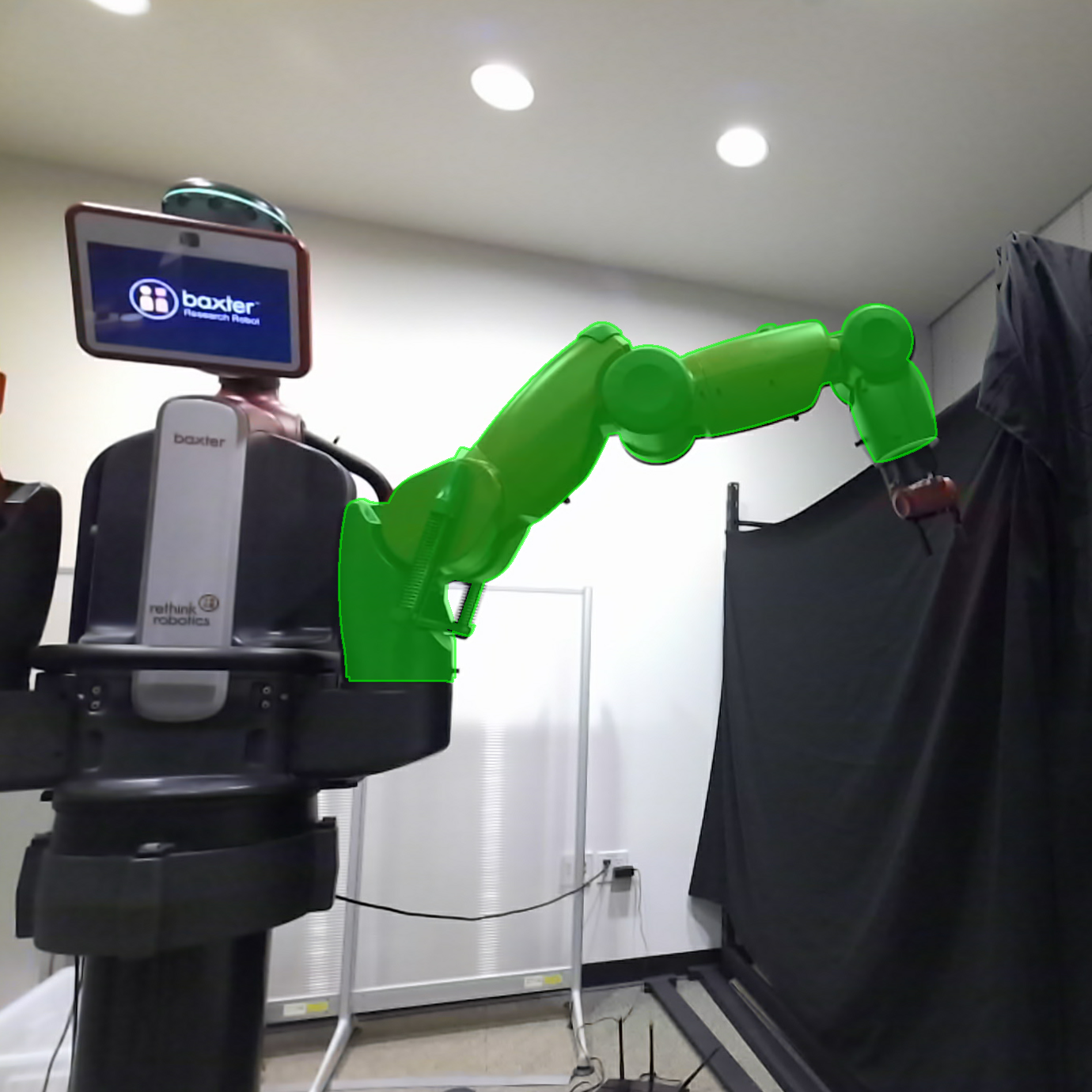} &
            \includegraphics[width=0.5\linewidth,trim={0cm 0cm 0cm 0cm},clip]{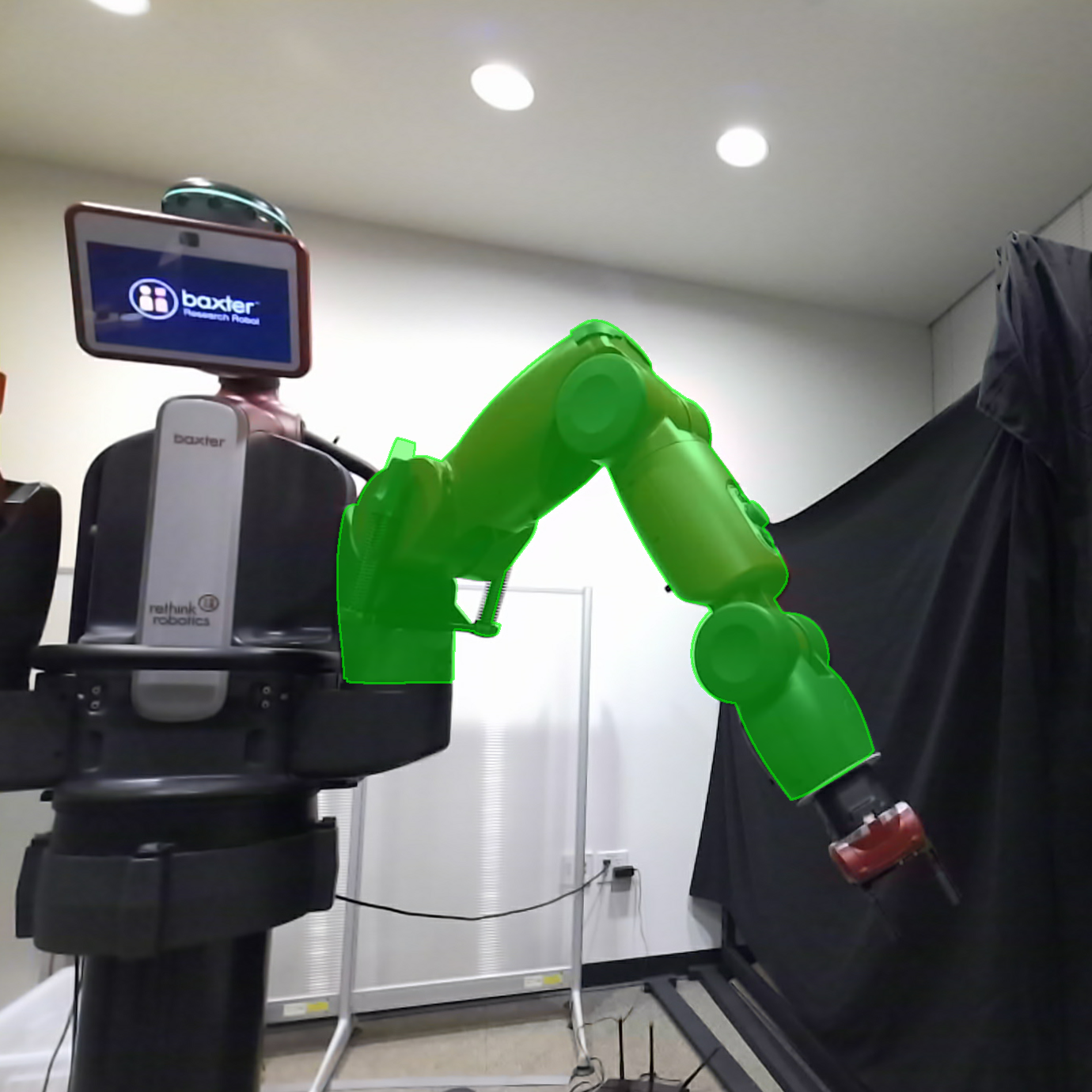}\\
            (a) & (b) \\
        \end{tabular}
    }
    \caption{\textbf{Qualitative results on the Baxter dataset.}
    We render the links of the Baxter robot arm with the solved camera poses and hover the masks on the RGB images.
    }
    \label{fig:baxter_result_example}
\end{figure}

%% file: tables/baxter_main_2d.tex
\begin{table}[ht]
    \centering
    \renewcommand{\arraystretch}{1.1}
    \resizebox{0.5\textwidth}{!}{
        \begin{tabular}{cccccccc}
            \hline
            Method & 20px   & 30px   & 40px   & 50px   & 100px  & 150px  & 200px  \\
            \Xhline{3\arrayrulewidth}
            DREAM~\cite{lee2020dream}  & 0.16 & 0.23 & 0.29 & 0.33 & 0.52 & 0.62 & 0.64 \\
            \hline
            OK~\cite{lu2022pose}     & 0.34 & 0.54 & 0.66 & 0.69 & 0.88 & 0.93 & 0.95 \\
            \hline
            IPE~\cite{lu2023image} (box) & - & - & - & 0.65 & 0.94 & 0.95 & 0.95 \\
            \hline
            IPE~\cite{lu2023image} (cylinder) & - & - & - & 0.80 & 0.91 & 0.93 & 0.95 \\
            \hline
            IPE~\cite{lu2023image} (CAD) & - & - & - & 0.74 & 0.90 & 0.94 & \textbf{1.00} \\
            \hline
            Ours (1view) & 0.35 & 0.55 & 0.75 & 0.90 & 0.95 & 0.95 & \textbf{1.00} \\
            \hline
            Ours (2views) & 0.40 & 0.75 & 0.95 & \textbf{1.00} & \textbf{1.00} & \textbf{1.00} & \textbf{1.00} \\
            \hline
            Ours (3views) & \textbf{0.55} & \textbf{0.85} & \textbf{1.00} & \textbf{1.00} & \textbf{1.00} & \textbf{1.00} & \textbf{1.00} \\
            \Xhline{3\arrayrulewidth}
        \end{tabular}
    }

    % \vspace{-0.3cm}
    \caption{
        \textbf{2D PCK evaluation results on the Baxter dataset.} \rev{We provide the 2D PCK scores at different thresholds, from 20 to 200 pixels.}
    }
    % \vspace{-3em}
    \label{tab:baxter_main_2d}
\end{table}

%% file: tables/baxter_main_3d.tex
\begin{table}[ht]
    \centering
    \renewcommand{\arraystretch}{1.1}
    \resizebox{0.5\textwidth}{!}{
        \begin{tabular}{ccccccc}
            \hline
            Method & 2cm & 5cm & 10cm & 20cm & 30cm & 40cm \\
            \Xhline{3\arrayrulewidth}
            DREAM~\cite{lee2020dream} & 0.01 & 0.08 & 0.32 & 0.43 & 0.54 & 0.66 \\
            \hline
            OK~\cite{lu2022pose} & 0.10 & 0.34 & 0.54 & 0.66 & 0.69 & 0.88 \\
            \hline
            IPE~\cite{lu2023image} (box) & - & - & 0.8 & 0.95 & 0.95 & 0.95 \\
            \hline
            IPE~\cite{lu2023image} (cylinder) & - & - & 0.71 & 0.93 & 0.94 & 0.95 \\
            \hline
            IPE~\cite{lu2023image} (CAD) & - & - & 0.78 & 0.93 & 0.97 & \textbf{1.00} \\
            \hline
            Ours (1view) & 0.10 & 0.65 & 0.90 & \textbf{1.00} & \textbf{1.00} & \textbf{1.00} \\
            \hline
            Ours (2views) & \textbf{0.15} & \textbf{0.80} & \textbf{0.95} & \textbf{1.00} & \textbf{1.00} & \textbf{1.00} \\
            \hline
            Ours (3views) & \textbf{0.15} & \textbf{0.80} & 0.90 & \textbf{1.00} & \textbf{1.00} & \textbf{1.00} \\
            \Xhline{3\arrayrulewidth}
        \end{tabular}
    }

    \caption{
        \textbf{3D PCK evaluation results on the Baxter dataset.} \rev{We provide the 3D PCK scores at different thresholds, from 2 to 40 cm.}
    }
   % \vspace{-3em}
    \label{tab:baxter_main_3d}
\end{table}

%% file: tables/xarm_real.tex
\begin{table}[!t]
    \centering
    \renewcommand{\arraystretch}{1.15}
    \resizebox{0.35\textwidth}{!}{
        \begin{tabular}{cccccccc}
            \hline
            Method     & DREAM & Ours & Ours (SAM) \\
            \Xhline{3\arrayrulewidth}
            Error (cm)      & 1.5   & 0.4  & 0.3        \\
            \Xhline{3\arrayrulewidth}
        \end{tabular}
    }

    % \vspace{-0.3cm}
    \caption{
        \rev{\textbf {Evaluation results in the real-world robot targeting experiments.}
        Ours uses PointRend for segmentation, and Ours (SAM) uses the Segment-Anything model with manually annotated bounding boxes as prompts.}
    }
    \label{tab:xarm_real}
\end{table}

%% file: figures/baxter_ablation.tex
\begin{figure}[bt]
    \centering
    \includegraphics[width=\linewidth]{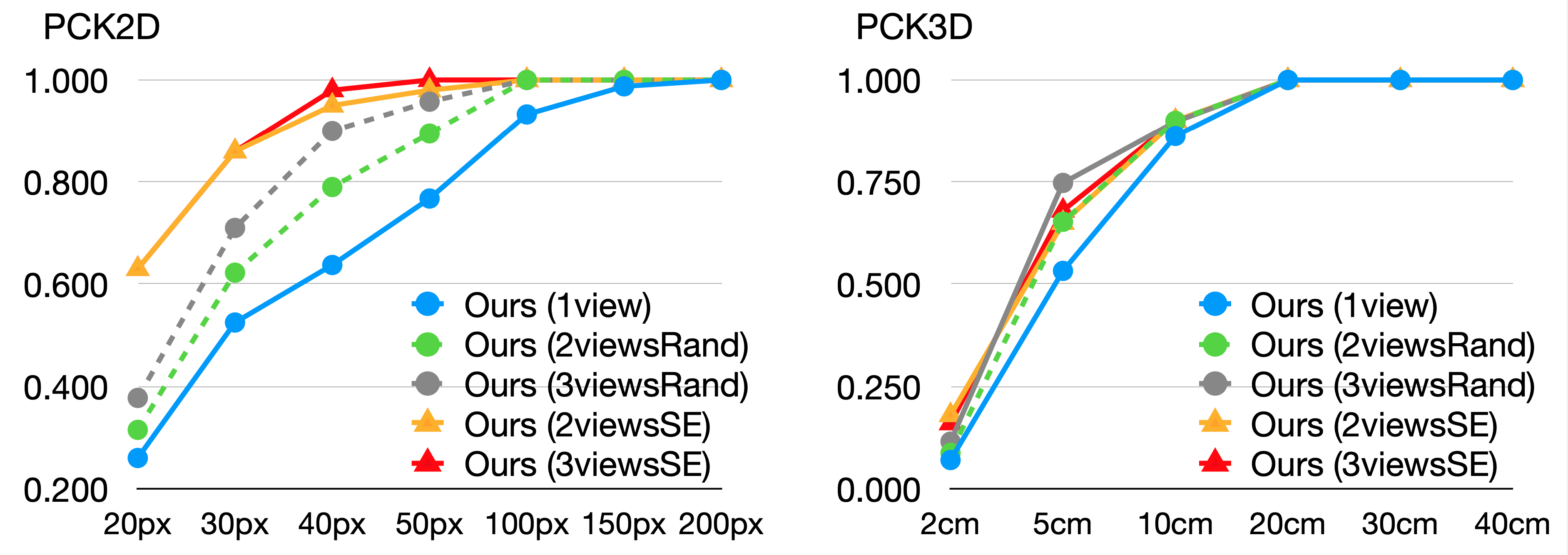}
%    \vspace{-1em}
    \caption{\textbf{Ablation study on the Baxter dataset.}
    We report the PCK2D (left) and PCK3D (right) on the Baxter dataset with different thresholds.
    Ours (SE) and Ours (Rand) represent our method with the next joint pose generation using consistency-based space exploration and random sampling, respectively.
    }
   % \vspace{-2em}
    \label{fig:baxter_ablation}
\end{figure}

%% file: figures/xarm_syn_ablation_trans.tex
\begin{figure}[bt]
    \includegraphics[width=\columnwidth]{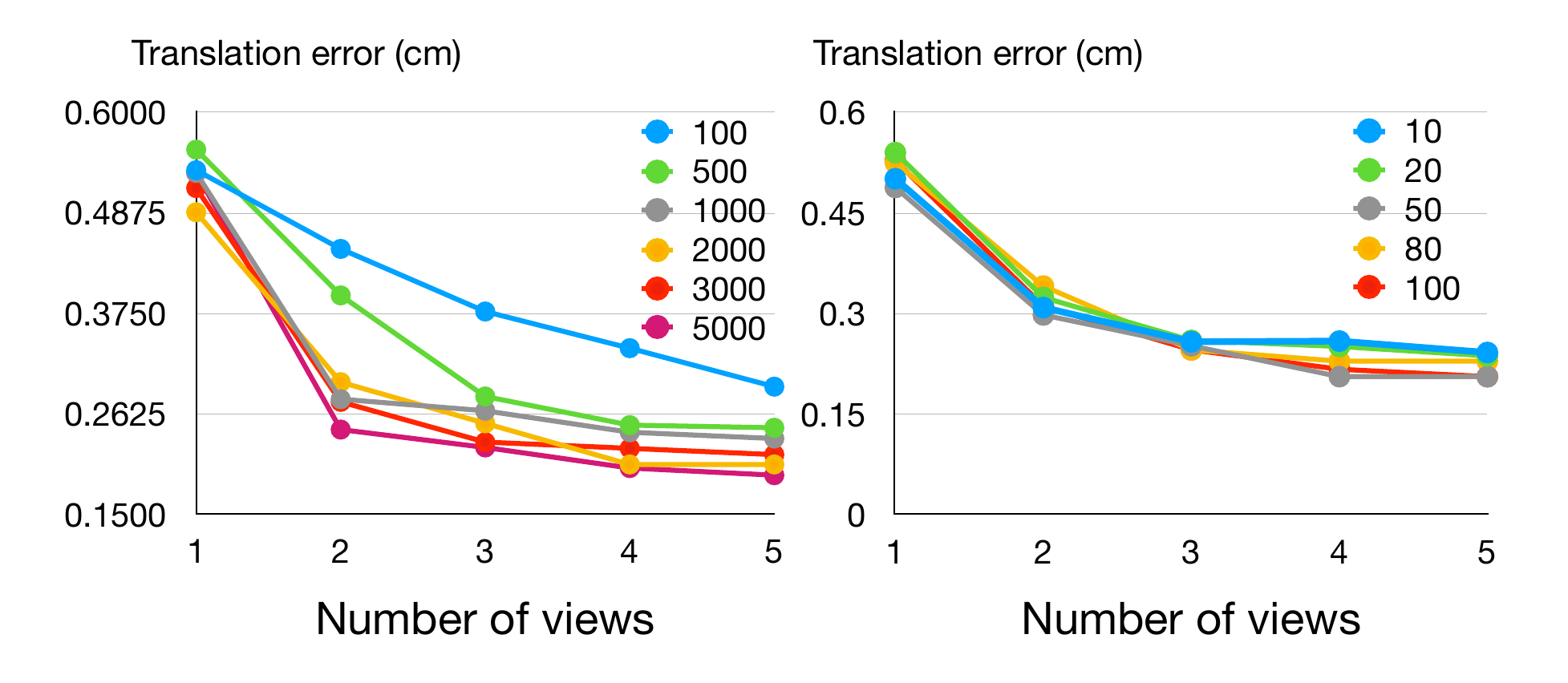}
    \caption{\rev{\textbf{Pose estimation results on the xArm synthetic dataset with different numbers of sampled joint poses (left) and camera pose candidates (right) in the joint space exploration process.}
    We report the translation errors in centimeters.}
    }
    \label{fig:xarm_syn_ablation_trans}
\end{figure}

%% file: tables/runtime.tex
\begin{table}[t]
    \centering
    \renewcommand{\arraystretch}{1.2}
    \resizebox{0.5\textwidth}{!}{
        \begin{tabular}{cccccccc}
            \hline
            Overall   & DR-based optim  & Space exploration & Robot movement \\
            \Xhline{3\arrayrulewidth}
             250-370s & 40-50s & 150-200s & 60-120s \\
            \Xhline{3\arrayrulewidth}
        \end{tabular}
    }

    \caption{
        \textbf {Runtime of EasyHeC tested on an RTX 4090 GPU.} We use 1000 optimization steps for the DR-based optimization, and 5 iterations for space exploration, with each iteration involving 2000 sampled joint poses and 50 camera pose candidates.
    }
    % \vspace{-1em}
    \label{tab:runtime}
\end{table}

%% file: figures/cotpc.tex
\begin{figure}[t]
    \centering
    \resizebox{1.0 \columnwidth}{!}{
        \begin{tabular}{cc}
            \includegraphics[width=0.5\linewidth,clip]{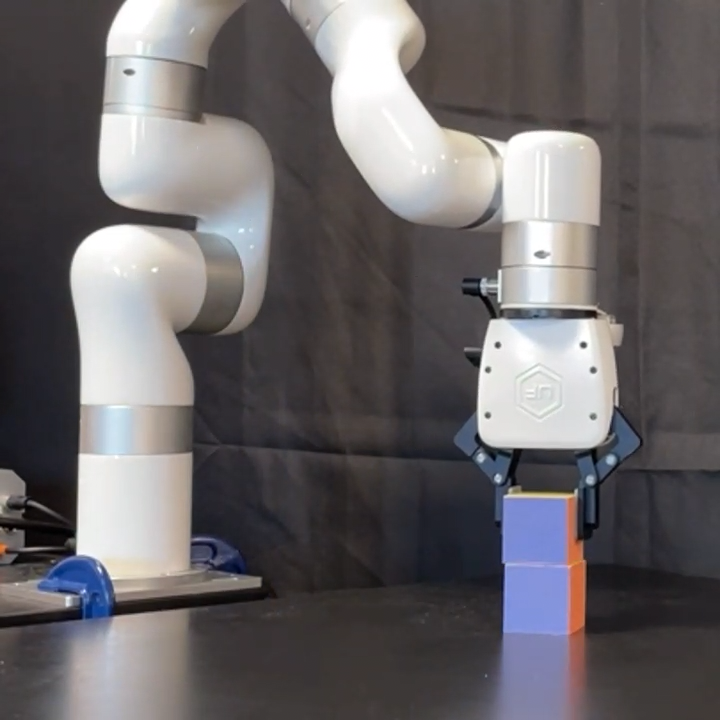} &
            \includegraphics[width=0.5\linewidth,clip]{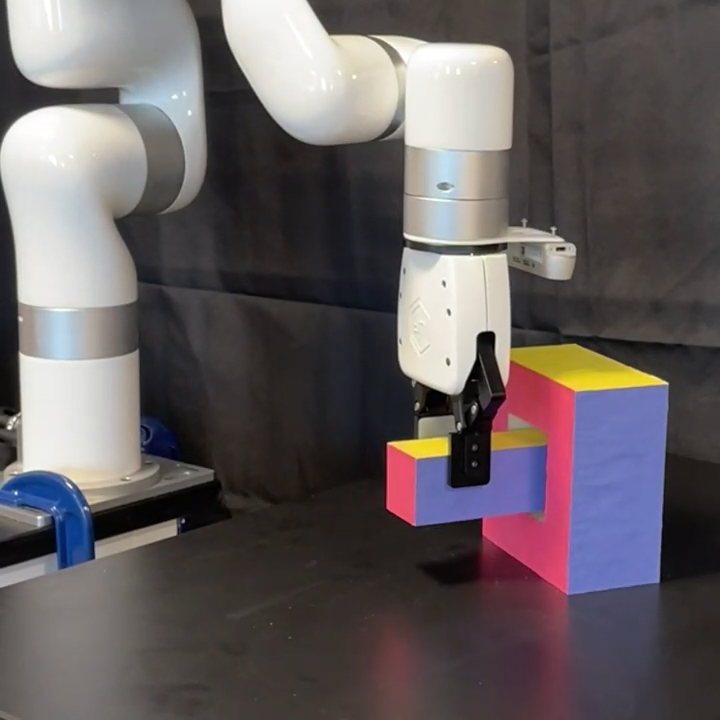}\\
            (a) & (b) \\
        \end{tabular}
    }
    \caption{\textbf{Real-world applications powered by our method:}
    screenshots from stacking cube (a) and peg insertion (b) experiments.
    Please refer to the accompanying supplementary video for a complete visualization of the process.
    }
    \label{fig:cotpc}
\end{figure}

%% file: sections/5_limitation.tex
\section{LIMITATION}\label{sec:limitation}

There are several limitations to our work.
\textbf{(1)} Our method necessitates a precise robot model, which may not enjoy universal applicability due to the presence of protective shells enveloping certain robots.
Nonetheless, the object shape of the robot links can be effectively optimized without ambiguity in the context of multiview calibration.
Alternatively, one can employ advanced techniques such as~\cite{yang2023movingparts} to reconstruct high-fidelity link shapes before using our method.
\textbf{(2)} Our method exclusively caters to the eye-to-hand calibration scenario.
However, for eye-in-hand configurations, the same pipeline can be leveraged for calibration.
Although the restricted field of view for the camera in this scenario may introduce inaccuracies when using the PVNet~\cite{peng2019pvnet} for initialization, the distribution range of the initial camera pose will be greatly constrained for eye-in-hand configurations, enabling manual initialization or the selection of one of several predefined camera poses for initialization.
\revsec{\textbf{(3)} PointRend needs to be retrained for a new robot arm. However, it’s a one-time training. Once trained, the model could be used for all the robot arms belonging to the same type. Moreover, we can also use the off-the-shelf Segment-Anything model for segmentation, with only several manually bounding box prompts annotation.}

%% file: sections/6_conclusions.tex
\section{CONCLUSIONS}
In this paper, we have presented a novel approach to hand-eye calibration.
Our method leverages the differentiable rendering technique to optimize the camera pose by minimizing the difference between the rendered mask and the observed mask.
The joint space exploration strategy is employed to find the informative joint poses for the camera pose optimization, which reduces the number of samples needed for achieving accurate results.
Our experimental results demonstrate that our approach outperforms traditional methods and several state-of-the-art methods on both synthetic and real-world datasets.
Overall, our approach provides a practical solution for camera pose estimation in robotics and can be applied to a wide range of applications, such as object manipulation, grasping, and pick-and-place tasks.
We believe that our work can reduce the sim-to-real gap and contribute to the development of more efficient and accurate robotic systems.

%% file: main.bbl
\begin{thebibliography}{10}
\providecommand{\url}[1]{#1}
\csname url@samestyle\endcsname
\providecommand{\newblock}{\relax}
\providecommand{\bibinfo}[2]{#2}
\providecommand{\BIBentrySTDinterwordspacing}{\spaceskip=0pt\relax}
\providecommand{\BIBentryALTinterwordstretchfactor}{4}
\providecommand{\BIBentryALTinterwordspacing}{\spaceskip=\fontdimen2\font plus
\BIBentryALTinterwordstretchfactor\fontdimen3\font minus \fontdimen4\font\relax}
\providecommand{\BIBforeignlanguage}[2]{{%
\expandafter\ifx\csname l@#1\endcsname\relax
\typeout{** WARNING: IEEEtran.bst: No hyphenation pattern has been}%
\typeout{** loaded for the language `#1'. Using the pattern for}%
\typeout{** the default language instead.}%
\else
\language=\csname l@#1\endcsname
\fi
#2}}
\providecommand{\BIBdecl}{\relax}
\BIBdecl

\bibitem{garrido2014automatic}
S.~Garrido-Jurado, R.~Mu{\~n}oz-Salinas, F.~J. Madrid-Cuevas, and M.~J. Mar{\'\i}n-Jim{\'e}nez, ``Automatic generation and detection of highly reliable fiducial markers under occlusion,'' \emph{Pattern Recognition}, vol.~47, no.~6, pp. 2280--2292, 2014.

\bibitem{ilonen2011robust}
J.~Ilonen and V.~Kyrki, ``Robust robot-camera calibration,'' in \emph{2011 15th International Conference on Advanced Robotics (ICAR)}.\hskip 1em plus 0.5em minus 0.4em\relax IEEE, 2011, pp. 67--74.

\bibitem{tsai1989new}
R.~Y. Tsai, R.~K. Lenz \emph{et~al.}, ``A new technique for fully autonomous and efficient 3 d robotics hand/eye calibration,'' \emph{IEEE Transactions on robotics and automation}, vol.~5, no.~3, pp. 345--358, 1989.

\bibitem{labbe2021single}
Y.~Labb{\'e}, J.~Carpentier, M.~Aubry, and J.~Sivic, ``Single-view robot pose and joint angle estimation via render \& compare,'' in \emph{Proceedings of the IEEE/CVF Conference on Computer Vision and Pattern Recognition}, 2021, pp. 1654--1663.

\bibitem{sefercik2023learning}
B.~C. Sefercik and B.~Akgun, ``Learning markerless robot-depth camera calibration and end-effector pose estimation,'' in \emph{Conference on Robot Learning}.\hskip 1em plus 0.5em minus 0.4em\relax PMLR, 2023, pp. 1586--1595.

\bibitem{bahadir2022deep}
O.~Bahadir, J.~P. Siebert, and G.~Aragon-Camarasa, ``A deep learning-based hand-eye calibration approach using a single reference point on a robot manipulator,'' in \emph{2022 IEEE International Conference on Robotics and Biomimetics (ROBIO)}.\hskip 1em plus 0.5em minus 0.4em\relax IEEE, 2022, pp. 1109--1114.

\bibitem{park1994robot}
F.~C. Park and B.~J. Martin, ``Robot sensor calibration: solving ax= xb on the euclidean group,'' \emph{IEEE Transactions on Robotics and Automation}, vol.~10, no.~5, pp. 717--721, 1994.

\bibitem{daniilidis1999hand}
K.~Daniilidis, ``Hand-eye calibration using dual quaternions,'' \emph{The International Journal of Robotics Research}, vol.~18, no.~3, pp. 286--298, 1999.

\bibitem{andreff2001robot}
N.~Andreff, R.~Horaud, and B.~Espiau, ``Robot hand-eye calibration using structure-from-motion,'' \emph{The International Journal of Robotics Research}, vol.~20, no.~3, pp. 228--248, 2001.

\bibitem{heller2011structure}
J.~Heller, M.~Havlena, A.~Sugimoto, and T.~Pajdla, ``Structure-from-motion based hand-eye calibration using l$_\infty$ minimization,'' in \emph{CVPR 2011}.\hskip 1em plus 0.5em minus 0.4em\relax IEEE, 2011, pp. 3497--3503.

\bibitem{zhi2017simultaneous}
X.~Zhi and S.~Schwertfeger, ``Simultaneous hand-eye calibration and reconstruction,'' in \emph{2017 IEEE/RSJ International Conference on Intelligent Robots and Systems (IROS)}.\hskip 1em plus 0.5em minus 0.4em\relax IEEE, 2017, pp. 1470--1477.

\bibitem{lee2020dream}
T.~E. Lee, J.~Tremblay, T.~To, J.~Cheng, T.~Mosier, O.~Kroemer, D.~Fox, and S.~Birchfield, ``Camera-to-robot pose estimation from a single image,'' in \emph{2020 IEEE International Conference on Robotics and Automation (ICRA)}.\hskip 1em plus 0.5em minus 0.4em\relax IEEE, 2020, pp. 9426--9432.

\bibitem{lepetit2009epnp}
V.~Lepetit, F.~Moreno-Noguer, and P.~Fua, ``Epnp: An accurate o (n) solution to the pnp problem,'' \emph{International journal of computer vision}, vol.~81, no.~2, pp. 155--166, 2009.

\bibitem{valassakis2022learning}
E.~Valassakis, K.~Dreczkowski, and E.~Johns, ``Learning eye-in-hand camera calibration from a single image,'' in \emph{Conference on Robot Learning}.\hskip 1em plus 0.5em minus 0.4em\relax PMLR, 2022, pp. 1336--1346.

\bibitem{besl1992icp}
P.~J. Besl and N.~D. McKay, ``Method for registration of 3-d shapes,'' in \emph{Sensor fusion IV: control paradigms and data structures}, vol. 1611.\hskip 1em plus 0.5em minus 0.4em\relax Spie, 1992, pp. 586--606.

\bibitem{lu2022pose}
J.~Lu, F.~Richter, and M.~C. Yip, ``Pose estimation for robot manipulators via keypoint optimization and sim-to-real transfer,'' \emph{IEEE Robotics and Automation Letters}, vol.~7, no.~2, pp. 4622--4629, 2022.

\bibitem{lu2023image}
J.~Lu, F.~Liu, C.~Girerd, and M.~C. Yip, ``Image-based pose estimation and shape reconstruction for robot manipulators and soft, continuum robots via differentiable rendering,'' \emph{IEEE International Conference on Robotics and Automation (ICRA)}, 2023.

\bibitem{zhang2023active}
X.~Zhang, Y.~Xi, Z.~Huang, L.~Zheng, H.~Huang, Y.~Xiong, and K.~Xu, ``Active hand-eye calibration via online accuracy-driven next-best-view selection,'' \emph{The Visual Computer}, vol.~39, no.~1, pp. 381--391, 2023.

\bibitem{yang2023next}
J.~Yang, J.~Rebello, and S.~L. Waslander, ``Next-best-view selection for robot eye-in-hand calibration,'' \emph{arXiv preprint arXiv:2303.06766}, 2023.

\bibitem{kirillov2020pointrend}
A.~Kirillov, Y.~Wu, K.~He, and R.~Girshick, ``Pointrend: Image segmentation as rendering,'' in \emph{Proceedings of the IEEE/CVF conference on computer vision and pattern recognition}, 2020, pp. 9799--9808.

\bibitem{kundu20183d}
A.~Kundu, Y.~Li, and J.~M. Rehg, ``3d-rcnn: Instance-level 3d object reconstruction via render-and-compare,'' in \emph{Proceedings of the IEEE conference on computer vision and pattern recognition}, 2018, pp. 3559--3568.

\bibitem{gabay1982minimizing}
D.~Gabay, ``Minimizing a differentiable function over a differential manifold,'' \emph{Journal of Optimization Theory and Applications}, vol.~37, pp. 177--219, 1982.

\bibitem{paszke2017automatic}
A.~Paszke, S.~Gross, S.~Chintala, G.~Chanan, E.~Yang, Z.~DeVito, Z.~Lin, A.~Desmaison, L.~Antiga, and A.~Lerer, ``Automatic differentiation in pytorch,'' 2017.

\bibitem{Xiang_2020_SAPIEN}
F.~Xiang, Y.~Qin, K.~Mo, Y.~Xia, H.~Zhu, F.~Liu, M.~Liu, H.~Jiang, Y.~Yuan, H.~Wang, L.~Yi, A.~X. Chang, L.~J. Guibas, and H.~Su, ``{SAPIEN}: A simulated part-based interactive environment,'' in \emph{The IEEE Conference on Computer Vision and Pattern Recognition (CVPR)}, June 2020.

\bibitem{peng2019pvnet}
S.~Peng, Y.~Liu, Q.~Huang, X.~Zhou, and H.~Bao, ``Pvnet: Pixel-wise voting network for 6dof pose estimation,'' in \emph{Proceedings of the IEEE/CVF Conference on Computer Vision and Pattern Recognition}, 2019, pp. 4561--4570.

\bibitem{laine2020modular}
S.~Laine, J.~Hellsten, T.~Karras, Y.~Seol, J.~Lehtinen, and T.~Aila, ``Modular primitives for high-performance differentiable rendering,'' \emph{ACM Transactions on Graphics (TOG)}, vol.~39, no.~6, pp. 1--14, 2020.

\bibitem{bradski2000opencv}
G.~Bradski, ``The opencv library.'' \emph{Dr. Dobb's Journal: Software Tools for the Professional Programmer}, vol.~25, no.~11, pp. 120--123, 2000.

\bibitem{kirillov2023segany}
A.~Kirillov, E.~Mintun, N.~Ravi, H.~Mao, C.~Rolland, L.~Gustafson, T.~Xiao, S.~Whitehead, A.~C. Berg, W.-Y. Lo, P.~Doll{\'a}r, and R.~Girshick, ``Segment anything,'' \emph{arXiv:2304.02643}, 2023.

\bibitem{jia2023chain}
Z.~Jia, F.~Liu, V.~Thumuluri, L.~Chen, Z.~Huang, and H.~Su, ``Chain-of-thought predictive control with behavior cloning,'' in \emph{Workshop on Reincarnating Reinforcement Learning at ICLR 2023}.

\bibitem{chen2023perceiving}
L.~Chen, Y.~Song, H.~Bao, and X.~Zhou, ``Perceiving unseen 3d objects by poking the objects,'' in \emph{2023 IEEE International Conference on Robotics and Automation (ICRA)}, 2023, pp. 4834--4841.

\bibitem{yang2023movingparts}
K.~Yang, X.~Zhang, Z.~Huang, X.~Chen, Z.~Xu, and H.~Su, ``Movingparts: Motion-based 3d part discovery in dynamic radiance field,'' \emph{Proceedings of the IEEE conference on computer vision and pattern recognition (CVPR)}, 2023.

\end{thebibliography}
